\documentclass[sigconf, nonacm]{acmart}

\usepackage{csquotes}
\usepackage{xcolor}
\usepackage{subcaption}
\usepackage{balance}
\colorlet{RED}{red}

\newcommand\vldbauthors{\authors}
\newcommand\vldbtitle{\shorttitle} 
\newcommand\vldbavailabilityurl{}
\newcommand\vldbpagestyle{plain}

\newcommand\system{ASET}

\definecolor{newteal}{HTML}{009988}
\definecolor{neworange}{HTML}{E58E26}
\definecolor{paradise_green}{HTML}{b8e994}
\definecolor{aurora_green}{HTML}{78e08f}
\definecolor{spray}{HTML}{82ccdd}
\definecolor{bright_yellow}{HTML}{fad390}
\definecolor{jalapeno_red}{HTML}{b71540}
\definecolor{forest_blue}{HTML}{0a3d62}
\begin{document}
\title{\system{}: \underline{A}d-hoc \underline{S}tructured \underline{E}xploration of \underline{T}ext Collections}
\subtitle{[Extended Abstract]}

%%
%% The "author" command and its associated commands are used to define the authors and their affiliations.
\author{Benjamin H{\"a}ttasch}
\orcid{0000-0001-8949-3611}
\affiliation{%
  \institution{TU Darmstadt}
}
%\email{benjamin.haettasch@cs.tu-darmstadt.de}
\authornote{Both authors contributed equally}

\author{Jan-Micha Bodensohn}
\affiliation{%
  \institution{TU Darmstadt}
}
%\email{jan-micha.bodensohn@stud.tu-darmstadt.de}
\authornotemark[1]

\author{Carsten Binnig}
\affiliation{%
  \institution{TU Darmstadt}
}
%\email{carsten.binnig@cs.tu-darmstadt.de}

\begin{abstract}
In this paper, we  propose a new system called \system{} that allows users to perform structured explorations of text collections in an ad-hoc manner.
The main idea of \system{} is to use a new two-phase approach that first extracts a superset of information nuggets from the texts using existing extractors such as named entity recognizers and then matches the extractions to a structured table definition as requested by the user based on embeddings.
In our evaluation, we show that \system{} is thus able to extract structured data from real-world text collections in high quality without the need to design extraction pipelines upfront.
\end{abstract}

\maketitle

\pagestyle{\vldbpagestyle}
%\begin{comment}
\begingroup\small\noindent\raggedright\textbf{AIDB Workshop Reference Format:}\\
\vldbauthors. \vldbtitle. AIDB 2021.
\endgroup
%\begingroup
%\end{comment}

%\vspace{-2em}

\begingroup
\renewcommand\thefootnote{}\footnote{

\noindent
This article is published under a Creative Commons Attributions License (http://creativecommons.org/licenses/by/3.0), which permits distribution and reproduction in any medium as well allowing derivative works, provided that you attribute the original work to the author(s) and AIDB 2021. 
\emph{3rd International Workshop on Applied AI for Database Systems and Applications (AIDB’21), August 20, 2021, Copenhagen, Denmark.}}\addtocounter{footnote}{-1}\endgroup

\ifdefempty{\vldbavailabilityurl}{}{
\vspace{.3cm}
\begingroup\small\noindent\raggedright\textbf{PVLDB Artifact Availability:}\\
The source code, data, and/or other artifacts have been made available at \url{\vldbavailabilityurl}.
\endgroup
}

\section{Introduction}

\begin{figure}
    \centering
    \includegraphics[width=.95\columnwidth]{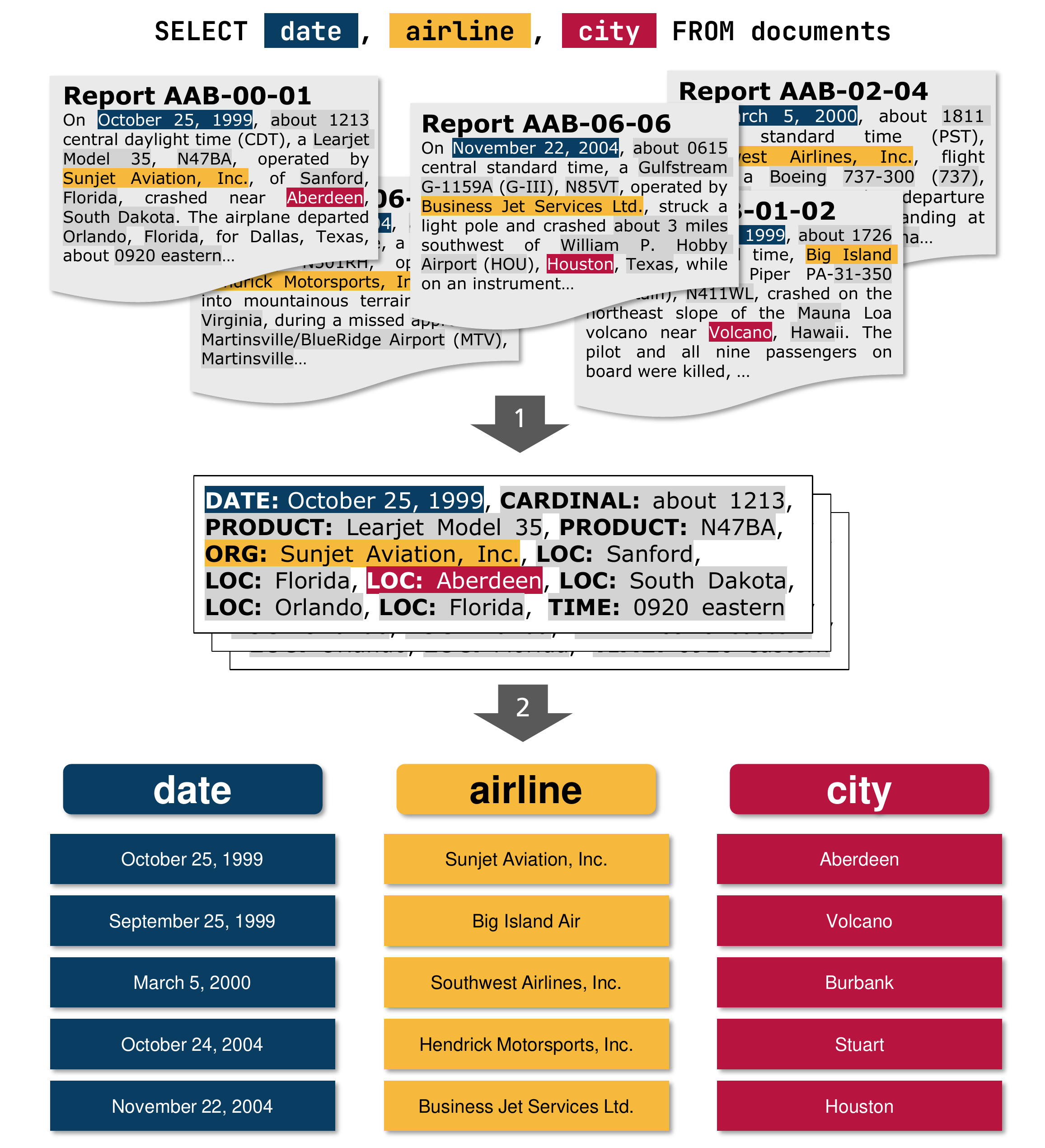}
    \caption{Ad-hoc structured exploration of text collections with \system{}: (1) a superset of information nuggets is first extracted from the texts and then (2) matched to the relevant attributes of the user query.}
    \vspace{-3.5ex}
    \label{fig:motivation}
\end{figure}

\paragraph*{Motivation.} In many domains, users face the problem of needing to quickly extract insights from large collections of textual documents. 
For example, imagine a journalist who wants to write an article about airline security that was triggered by some recent incidents of a well-known US airline.
For this reason, the journalist might decide to explore a collection of textual accident reports from the \textit{National Transportation Safety Board} in order to answer questions like 'What incident types are the most frequent ones?' or 'Which airlines are involved most often in incidents?'. 
To be able to formulate such answers to their questions, they would need to extract the relevant information, then create a structured data set (e.g., by creating a table in a database or simply by using an Excel sheet) and analyze frequency statistics such as the number of incidents per airline.

And clearly, there are many more domains where end users want to explore textual document collections in a similar fashion.
As another example, think of medical doctors who want to compare symptoms and reactions to medical treatments for different groups of patients (e.g., old vs. young, w/ or w/o a specific pre-existing condition) based on the available data coming from textual patient reports.
To do this, the doctor again would need to extract the relevant structured information about age, pre-diseases, etc. from those reports before being able to draw any conclusions.

One could now argue that extracting structured data from text is a classical problem that various communities have already tackled and several industry-scale systems already exist.
For example, DeepDive \cite{DeepDive-2016-Sa} or System-T \cite{lembo2020ontology} are examples of such systems that have developed rather versatile tool suites to extract structured facts from textual sources.
However, these systems typically require a team of highly-skilled engineers that curate extraction pipelines to populate a structured database from the given text collection or train machine learning-based extraction models that come with the additional need to curate labeled training data.
A major problem of these solutions is the high effort they require and, thus, it can take days or weeks to curate such extraction pipelines even if experts are involved.
Even more importantly, such extraction pipelines are typically rather static and can extract only a pre-defined (i.e., fixed) set of attributes for a certain text collection only. 
This typically prevents more exploratory scenarios in which users ask ad-hoc queries where it is not known upfront which information needs to be extracted or whether a new data set should be supported on-the-fly.

\begin{figure*}
    \centering
    \includegraphics[width=\textwidth]{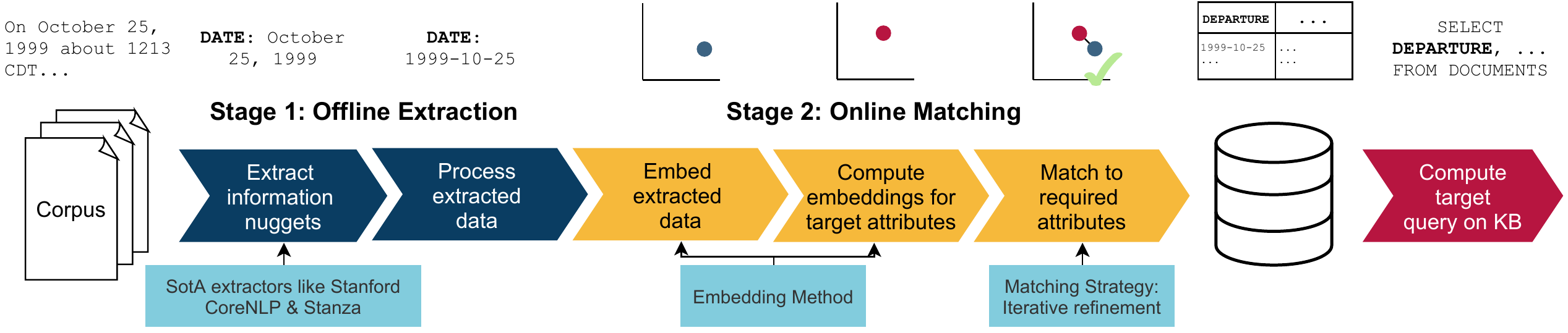}
    \caption{Architecture of \system{}: The extraction stage obtains information nuggets from the documents. The matching stage matches between the extracted information nuggets and the user's schema imposed by their query.}
    \label{fig:overview}
\end{figure*}

\vspace{-1.5ex}
\paragraph*{Contributions.}  In this paper, we thus propose a new system called \system{} that allows users to explore new (unseen) text collections by deriving structured data in an ad-hoc manner; i.e., without the need to curate extraction pipelines for this collection.

As shown in Figure \ref{fig:motivation}, the main idea is that a user specifies their information need by composing SQL-style queries over the text collection. 
For example, in Figure \ref{fig:motivation}, the user issues a query to extract information about dates, airlines, and cities of incidents. 
\system{} then takes the query and evaluates it over the given document collection by automatically populating the table(s) required to answer the query with information nuggets from the documents.
An important aspect here is that using \system{}, users can define their information needs by such a query in an ad-hoc manner.

\system{} supports this ad-hoc extraction of structured information by implementing a new two-phase approach:
In the \emph{extraction phase}, a superset of information nuggets is extracted from a text collection. 
Afterward, the information nuggets are matched to the required attributes in the \emph{matching phase}.
To do so, \system{} implements a new interactive approach for matching based on neural embeddings which uses a tree-based exploration technique to identify potential matches for each attribute.

Clearly, doing the matching for arbitrarily complex document collections and user queries is a challenging task.
Hence, in this paper we aim to show a first feasibility study of our approach.
To that end, we focus on so-called \emph{topic-focused} document collections here. 
In these collections every document provides the same type of information (e.g., an airline incident) meaning that each document can be mapped to one row of a single extracted table.
Note that this is still a challenging task since arbitrary information nuggets must be mapped to an extracted table in an ad-hoc manner.
Clearly, extending this to more general document collections and queries that involve multiple tables is an interesting avenue of future work.

To summarize, as the main contribution in this paper we present the initial results of \system{}.
This comprises a description of our approach in Sections \ref{sec:overview} and \ref{sec:interactive_matching}  as well as an initial evaluation in Section \ref{sec:evaluation} on two real-world data sets.
In addition, we provide code and data sets for download\footnote{\url{https://github.com/DataManagementLab/ASET}} along with a short video of \system{}.$\!$\footnote{\url{https://youtu.be/IzcT8kn8bUY}}

\section{Overview of our Approach}% \todo{(1 Page)}
\label{sec:overview}

Figure \ref{fig:overview} shows the architecture of \system{}.
As mentioned before, \system{} comprises two stages:
(1) the first stage extracts a superset of potential information nuggets from a collection of input documents using extractors. 
This step is independent of the user queries and can thus be executed offline to prepare the text collection for ad-hoc exploration by the user.
(2) At runtime, a user issues several queries against \system{}. 
To answer a query, an online matching stage is executed that aims to map the information nuggets extracted in the first stage to the attributes of the user table as requested by a query.
We make use of existing state-of-the-art approaches for information extraction in the first stage.
The core contribution of \system{} is in the matching stage which implements a novel tree-based approach as we discuss below.

\subsection{Stage 1: Offline Extraction}

As shown in Figure \ref{fig:overview}, the extraction stage is composed of two steps. First, it derives the information nuggets as label-mention-pairs (e.g., a date and its textual representation) from the source documents using state-of-the-art extractors. 
Afterward, a preprocessing step is applied which then canonicalizes the extracted values. 

\vspace{-1.5ex}
\paragraph*{Extracting Information Nuggets.} The extractors process the collection document by document
to generate the corresponding extractions. 
Clearly, a limiting factor of \system{} is which kinds of information nuggets can be extracted in the extraction stage since only this information can be used for the subsequent matching stage.
For this paper, we successfully employed state-of-the-art information extraction systems, focusing particularly on named entity recognizers from \textit{Stanford CoreNLP} \cite{Stanford-CoreNLP-2014-Manning} and \textit{Stanza} \cite{Stanza-2020-Qi}. 
In general, \system{} can be used with any extractor that produces label-mention pairs;
i.e. a textual mention of a type in the text (e.g., \emph{American Airlines}) together with a label representing its semantic type (e.g., \emph{Company}).
Moreover, additional information about the extraction (e.g., the full sentence around the mention) will also be stored and used for computing the embeddings as we describe below.

\vspace{-1.5ex}
\paragraph*{Preprocessing Extracted Data.} In the last step of the extraction stage, the extractions are preprocessed to derive their actual data values from their mentions (i.e. a canonical representation). 
For this we also rely on state-of-the-art systems for normalization:
As an example, we employ Stanford CoreNLP's \cite{Stanford-CoreNLP-2014-Manning} built-in, rule-based temporal expression recognizer SUTime for normalization of dates (e.g., turn \emph{October 25, 1999} and \emph{25.10.1999} into \emph{1999-10-25}).

\subsection{Stage 2: Online Matching}

The second stage must match the extracted information nuggets to the user table to answer the query. 
This stage consists of computing embeddings for the information nuggets and matching them to the target attributes using a new tree-based technique to identify groups of similar objects in the joint embedding space of attributes and information nuggets.

\vspace{-1.5ex}
\paragraph*{Computing Embeddings.}
A classical approach to compute a mapping between information nuggets and attributes of the user table would be to train a machine learning model in a supervised fashion. 
However, this would require both training time and a substantial set of labeled training data for each attribute and domain.
Instead, our approach leverages embeddings to quantify the intuitive semantic closeness between information nuggets and the attributes of the user table.$\!$\footnote{We use Sentence-BERT \cite{Sentence-BERT-2019-Reimers} and FastText \cite{mikolov2018advances} to compute embeddings for the natural language signals.}
For the attributes of the user table, only the attribute names are available to derive an embedding. 
To embed the information nuggets extracted in the first stage, however, we can use more information and incorporate the following signals from the extraction:
(1) \emph{label} -- the entity type determined by the information extractor (e.g. \textit{Company}),$\!$\footnote{We map the named entity recognizers' labels like \textit{ORG} to suitable natural language expressions according to the descriptions in their specification.} 
(2) \emph{mention} -- the textual representation of the entity in the text (e.g., \textit{US Airways}), 
(3) \emph{context} -- the sentence in which the mention appears, 
(4) \emph{position} -- the position of the mention in the document.

\vspace{-1.5ex}
\paragraph*{Matching Step.} 
To populate the values of a row (i.e. to decide whether an extraction is a match for an attribute of a user table, like a specific \text{DATE} instance matches \text{DEPARTURE} in Figure \ref{fig:overview}), we use a new tree-based technique to identify groups of related information nuggets that map to a user attribute as we discuss in the next section.
Using this technique, we suggest potential matches to the user who can confirm or reject those matches in an interactive manner (i.e., by reading the extracted value and its context sentence). 
Previous approaches use only a distance metric (e.g., cosine distance) and often suffer from the curse of dimensionality, not providing a robust similarity metric in higher-dimensional embedding spaces.
Our approach allows users to quickly explore the embedding space and find matches between extracted information nuggets and attributes more efficiently.

\begin{figure}
    \centering
    \includegraphics[width=\columnwidth]{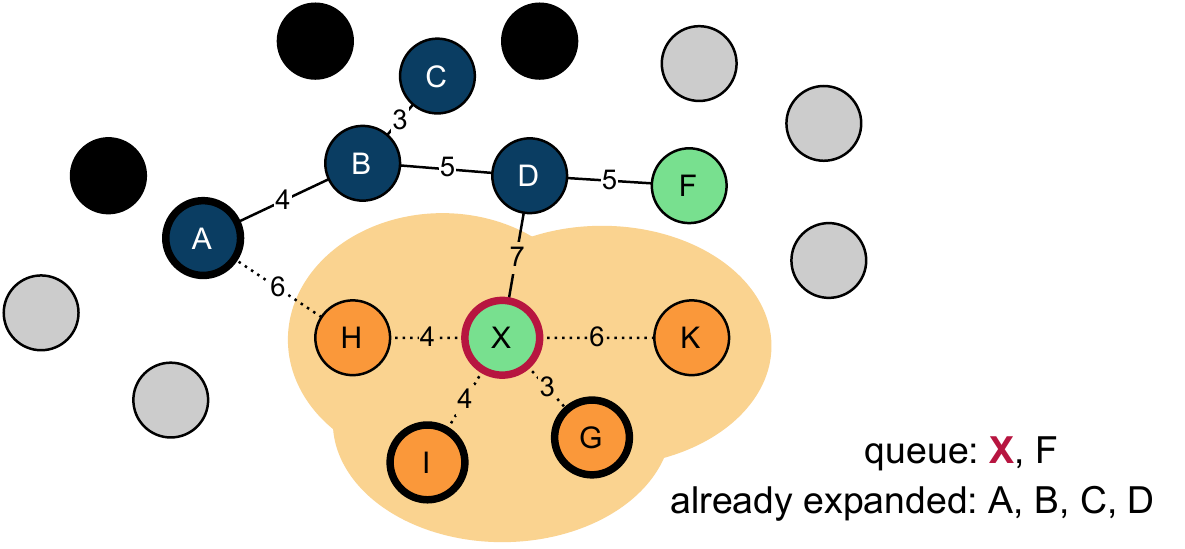}
    \caption{Sketch of the tree-based (explore-away) strategy (executed per attribute). Each node represents an embedded information nugget to be included in the search tree.  Confirmed matches are marked in \textcolor{forest_blue}{$\blacksquare$} blue, the next candidates in \textcolor{aurora_green}{$\blacksquare$} green. Rejected nuggets are marked in $\blacksquare$ black, unexplored ones in \textcolor{gray}{$\blacksquare$} gray. Node X is selected for expansion, the nodes closest to X are marked in \textcolor{neworange}{$\blacksquare$} orange. The candidates selected by our explore-away strategy for user-feedback are G and I. 
    }
    \label{fig:tree_search}
\end{figure}

\section{Interactive Matching} %\todo{(1 pages)}
\label{sec:interactive_matching}

In this section, we first give an overview of the interactive matching process before we discuss the details of how \system{} selects potential matches to present to the user.
The matching is done individually for the different attributes.

\subsection{Overall Procedure}

\system{} implements an interactive matching procedure by confronting the user with information nuggets derived from the document collection and asking them whether those nuggets belong to a particular attribute.
The main goal of the interactive matching procedure is to identify groups of information nuggets in the embedding space belonging to a particular user-requested attribute (e.g., airline names or incident types) as quickly as possible (i.e. after a low number of interactions).
However, finding information nuggets to present to the user as potential matches is not trivial.
Clearly, a first naive idea is to choose extractions that are close to the embedding of the requested attribute (e.g., airline name) using a distance metric (e.g., cosine similarity).
Yet, matches identified by these simple distances provide only a limited information gain for identifying additional groups of related objects in a high-dimensional space.
Therefore, we are using a new tree-based exploration strategy that  can efficiently identify potential matches for each user-requested attribute as we discuss next.

\subsection{Tree-based Exploration Strategy}

\begin{figure*}
    \centering
    \begin{subfigure}[b]{0.49\textwidth}
        \centering
        \includegraphics[width=\columnwidth]{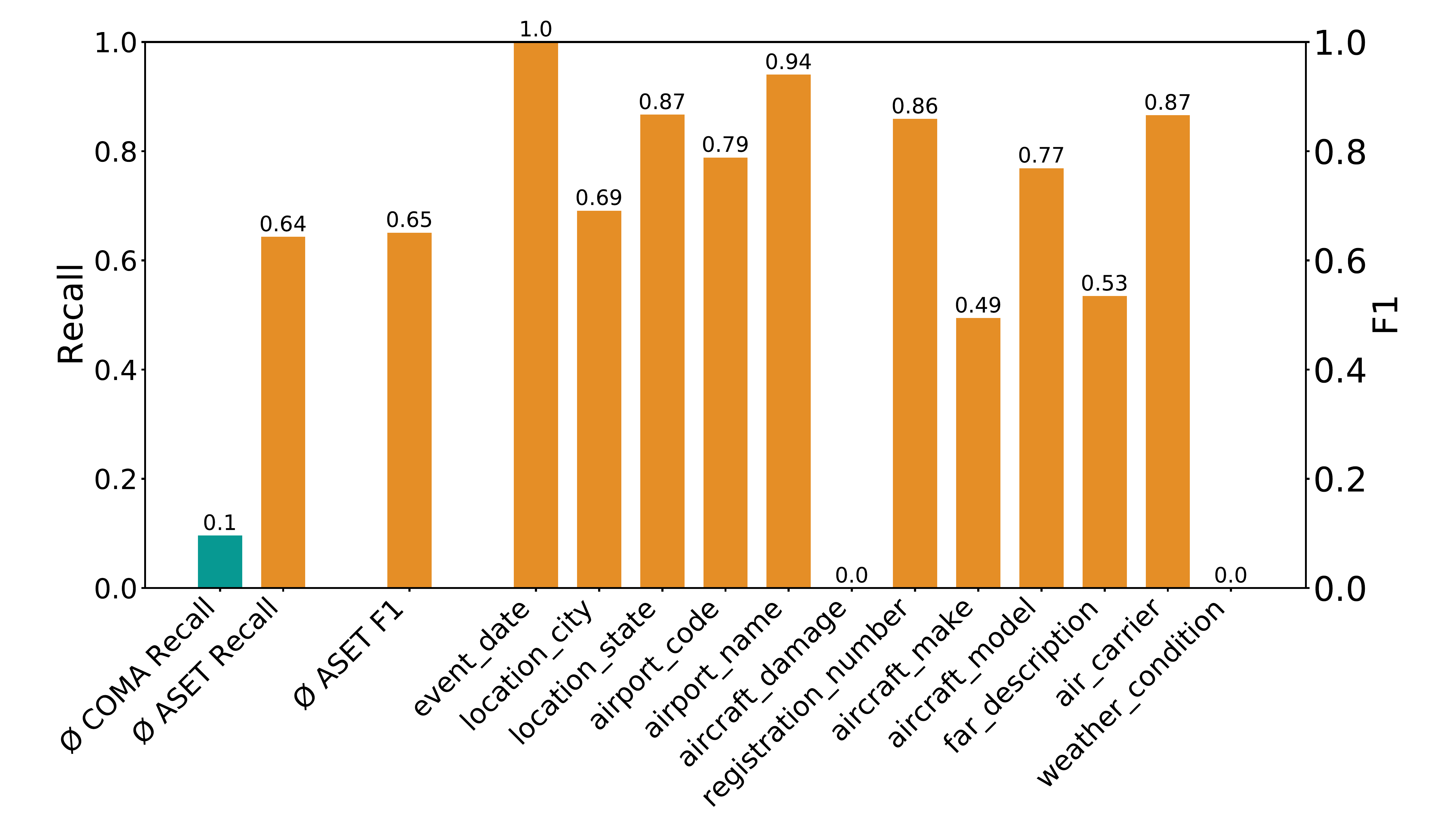}
    \end{subfigure}
    \begin{subfigure}[b]{0.49\textwidth}
        \centering
        \includegraphics[width=\columnwidth]{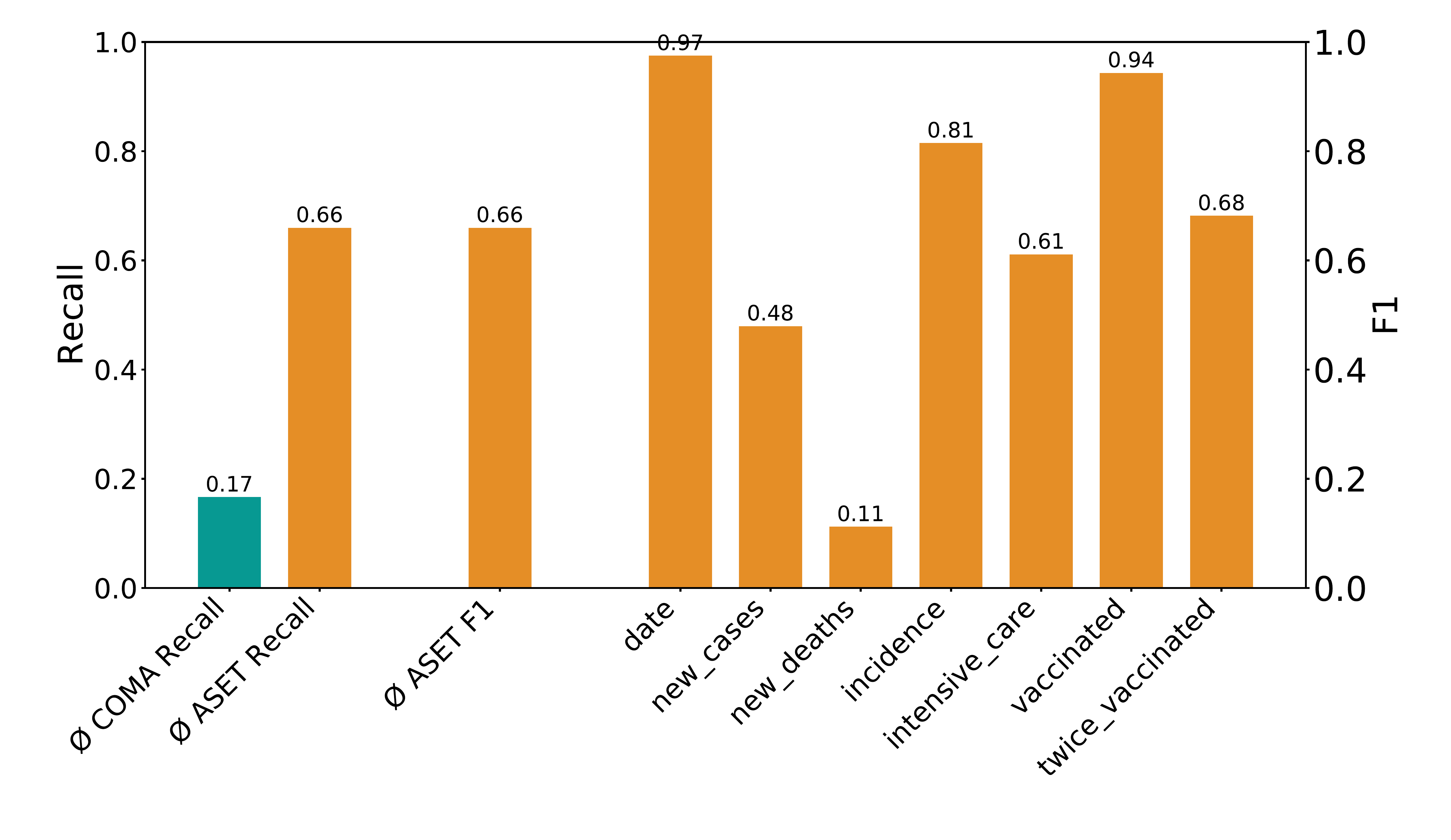}
    \end{subfigure}
    
    \vspace*{-2.5ex}
    \caption{End-to-end evaluation (both stages) of our system on both the \emph{Aviation} and the \emph{COVID19RKI} data set. We report the avg. F1-score and the F1-scores for all attributes for \system{} (all scores ranging from 0 to 1, higher is better).}
    \label{fig:exp2}
\end{figure*}

To identify the group of information nuggets in the embedding space that belong to a user-requested attribute, we use the notion of a tree of confirmed matching extractions (instead of a set as often used by kNN-based approaches). 
Different from kNN-searches, subspace clustering, or other techniques tackling similar problems, using a tree-based representation allows us to implement a new \emph{explore-away} strategy that can grow the covered embedding space for the group of related information nuggets with every confirmed match.
Our tree-based exploration strategy works in three steps:

\vspace{-1.5ex}
\paragraph*{1. Find a root node:}
First, the exploration strategy finds an initial matching node to serve as the root of the tree.
This is done by sampling extractions based on their distance to the initial attribute embedding (based only on the attribute name).
We start with low distances that result in conservative samples close to the initial attribute embedding and gradually raise the sampling temperature to include samples from farther away if the close-by samples do not yield any matching extractions to select as root.

\vspace{-1.5ex}
\paragraph*{2. Explore-away Expansion:}
As a second step, we now explore the embedding space by expanding the search tree using our explore-away strategy in the embedding space.
We explain the expansion step based on the example in Figure \ref{fig:tree_search} where node \textit{X} is to be expanded.
To expand the node \textit{X}, we determine its potential successors \texttt{succ(\textit{X})} based on the following two constraints:
(1) The extractions in \texttt{succ(\textit{X})} must be closer to \textit{X} than to any other already expanded extraction (e.g.,  nodes \textit{G}, \textit{H}, \textit{I}, and \textit{K} qualify in our example).
(2) The extractions in \texttt{succ(\textit{X})} must be farther away from the rest of the tree than the node we expand (e.g., \textit{H} is closer to \textit{A} than \textit{X} is to its parent (and hence closest node) \textit{D} and therefore not a candidate; however, nodes \textit{G}, \textit{I}, and \textit{K} remain as candidates).

Afterward, the search strategy selects the $k$ nuggets\footnote{This number determines the degree of the search trees. We experimented with different degrees and found that $2$ results in the best performance.} in \texttt{succ(\textit{X})} that are closest to \textit{X} (e.g., \textit{G} and \textit{I} in our example) to gather user feedback.
A user can then confirm whether the proposed nuggets actually match the attribute; matching nuggets are added to a queue of nuggets to be expanded in the next iterations.
In case the queue is empty, the explore-away strategy returns to step 1 to start with an additional root node or it terminates if a user-defined threshold of confirmed matches is reached.
Overall, this procedure thus identifies groups of information nuggets (represented as trees) that match to a certain user-requested attribute.

\vspace{-1.5ex}
\paragraph*{3. Static Matching:}
Once a user-defined threshold of confirmed matches is reached for every user attribute, \system{} stops collecting feedback and continues with a static matching procedure:
\system{} leverages the distances between the embeddings of extracted information nuggets and the different groups of embeddings identified in step 2 to populate the remaining cells without asking the user for feedback.

\section{Initial Evaluation} %\todo{(3 pages)}
\label{sec:evaluation}

In this section, we present an initial evaluation of our approach showing that \system{} can provide high accuracies.
For the evaluation, we use two different data sets, which we provide for download together with our source code.
We also have additional results showing that our novel tree-based exploration technique is superior over using other techniques that are only based on distance metrics (e.g., cosine similarity) in the  embedding space. 
However, due to space restrictions we could not include these results in this paper.

\vspace{-1.5ex}
\paragraph*{Data Sets}
We perform our evaluation on the two real-world data sets: \emph{Aviation} and \emph{COVID19RKI}.
Both data sets consist of a document collection and structured tables to serve as ground truth.

The \emph{Aviation} data set is based on the executive summaries of the Aviation Accident Reports published by the United States National Transportation Safety Board (NTSB).$\!$\footnote{\url{https://www.ntsb.gov/investigations/AccidentReports/Pages/aviation.aspx}}
Each report describes an aviation accident and provides details like the prevailing circumstances, probable causes, conclusions, and recommendations.
As a ground-truth we compiled a list of 12 typical attributes and manually created annotations that capture where the summaries mention the attributes' values. 

The second data set is based on the German RKI's daily reports outlining the current situation of the Covid-19 pandemic (e.g., laboratory-confirmed Covid-19 cases or the number of patients in intensive care) in Germany.$\!$\footnote{\url{https://www.rki.de/DE/Content/InfAZ/N/Neuartiges_Coronavirus/Situationsberichte/Gesamt.html}}
The focus of this data set is to evaluate whether our system can also cope with numerical values which are particularly challenging for matching. 
To that end, we compiled a list of seven numeric attributes and manually annotated the occurrences of those attributes in the data set.

\vspace{-1.5ex}
\paragraph*{Initial Results}

In this initial experiment, we evaluate the end-to-end performance of our system. 
As a baseline to compare the quality of the matching stage of \system{} to, we use COMA 3.0 \cite{do2002coma} which implements a wide set of classical matching strategies that also work out-of-the-box (i.e., similar to \system{} they do not need to be trained for every new attribute to be matched).

Figure \ref{fig:exp2} shows the result of running \system{} (using \emph{Stanza} in the extraction stage) with 25 interactions per attribute in the matching stage.
We report the average F1-score as well as the individual F1-scores for all attributes. 
The results show that \system{} is able to accurately match the extractions for most of the attributes of both data sets.
For COMA 3.0, we decided to report only the recall since the precision of matching values was overall low (depending on the selected workflows it either finds hardly any matches or thousands of wrong matches).

\section{Conclusions and Future Work}
In this paper, we have proposed a new system called \system{} for ad-hoc structured exploration of text collections. 
Overall, we have shown that \system{} is able to extract structured data from real-world text collections in high quality without the need to  manually curate extraction pipelines.
In the future, we plan to extend our system in several directions; e.g., to support more complex user queries (e.g., with joins over multiple tables) or more complex document collections.

\begin{acks}
This work has been supported by the German Federal Ministry of Education and Research as part of the Project Software Campus 2.0 (TUDA), Microproject INTEXPLORE, under grant ZN 01IS17050, as well as by the German Research Foundation as part of the Research Training Group \textit{Adaptive Preparation of Information from Heterogeneous Sources} (AIPHES) under grant No. GRK 1994/1, and through the project NHR4CES. 
\end{acks}

%\clearpage
%\balance{}
\bibliographystyle{ACM-Reference-Format}
\bibliography{main}

\end{document}